%% file: acl_latex.tex
\pdfoutput=1

\documentclass[11pt]{article}

\usepackage[final]{acl}

\usepackage{times}
\usepackage{latexsym}

\usepackage[T1]{fontenc}

\usepackage[utf8]{inputenc}

\usepackage{microtype}

\usepackage{inconsolata}

\usepackage{graphicx}

\title{Anticipating Future with Large Language Model for Simultaneous Machine Translation}

\author{
 \textbf{Siqi Ouyang\textsuperscript{1}},
 \textbf{Oleksii Hrinchuk\textsuperscript{2}},
 \textbf{Zhehuai Chen\textsuperscript{2}},
 \textbf{Vitaly Lavrukhin\textsuperscript{2}},
\\
 \textbf{Jagadeesh Balam\textsuperscript{2}},
 \textbf{Lei Li\textsuperscript{1}},
 \textbf{Boris Ginsburg\textsuperscript{2}},
\\
 \textsuperscript{1}Carnegie Mellon University,
 \textsuperscript{2}NVIDIA,
\\
 \texttt{siqiouya@andrew.cmu.edu}
}

\usepackage{multicol}
\usepackage{multirow}
\usepackage{booktabs}
\usepackage{enumitem}
\input{math}

\newcommand{\method}{TAF}

\begin{document}
\maketitle

\input{latex/sections/0_abstract}
\input{latex/sections/1_introduction}
\input{latex/sections/2_related_works}
\input{latex/sections/3_method}
\input{latex/sections/4_experiment}
\input{latex/sections/5_analysis}
\input{latex/sections/6_conclusion}
\input{latex/sections/limitations}

\bibliography{clean}

\appendix
\input{latex/sections/appendix}

\end{document}

%% file: math.tex
\usepackage{amsmath,amsfonts,bm}

\def\eqref#1{equation~\ref{#1}}

\def\1{\bm{1}}

\def\vg{{\bm{g}}}

\def\vx{{\bm{x}}}
\def\vy{{\bm{y}}}

\DeclareMathAlphabet{\mathsfit}{\encodingdefault}{\sfdefault}{m}{sl}
\SetMathAlphabet{\mathsfit}{bold}{\encodingdefault}{\sfdefault}{bx}{n}

\newcommand{\E}{\mathbb{E}}



%% file: latex/sections/0_abstract.tex
\begin{abstract}

Simultaneous machine translation (SMT) takes streaming input utterances and incrementally produces target text.
Existing SMT methods mainly use the partial utterance that has already arrived at the input and the generated hypothesis.
Motivated by human interpreters' technique to forecast future words before hearing them, we propose \textbf{T}ranslation by \textbf{A}nticipating \textbf{F}uture (\method), a method to improve translation quality while retaining low latency. 
Its core idea is to use a large language model (LLM) to predict future source words and opportunistically translate without introducing too much risk.
We evaluate our \method~and multiple baselines of SMT on four language directions. Experiments show that \method~achieves the best translation quality-latency trade-off and outperforms the baselines by up to 5 BLEU points at the same latency (three words)\footnote{\url{https://github.com/owaski/TAF}}. 


\end{abstract}

%% file: latex/sections/1_introduction.tex
\section{Introduction}

\begin{figure}[ht]
    \centering
    \includegraphics[width=0.8\linewidth]{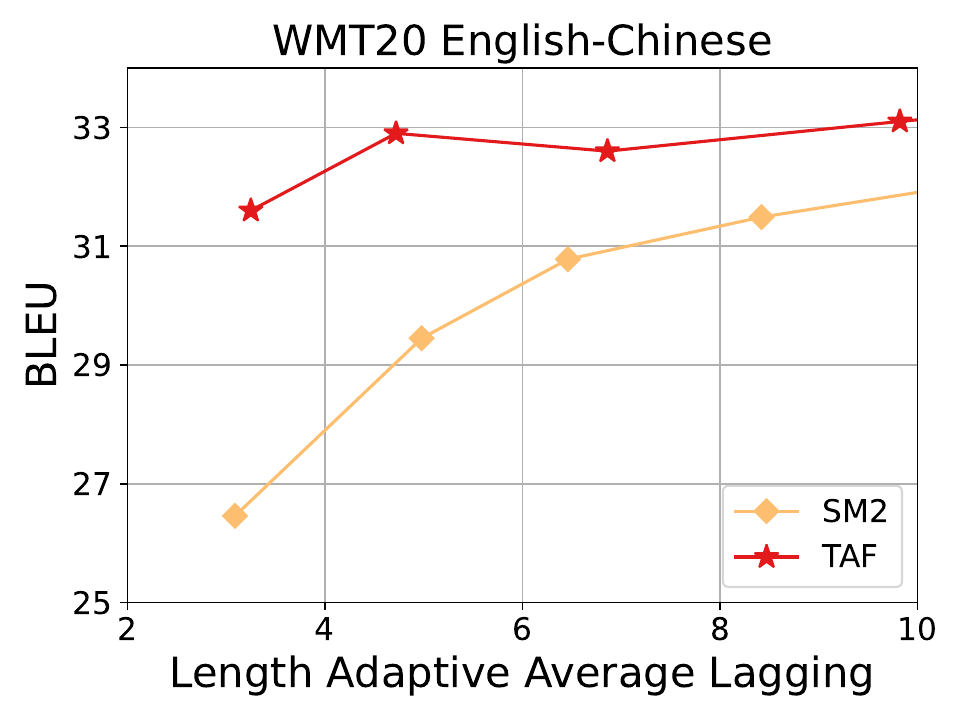}
    \caption{The quality-latency trade-off of our method \method~and the previous state-of-the-art method SM2 on WMT20 English-Chinese direction. The translation quality of SM2 drops quickly as latency goes down while \method~maintains a good translation quality even at the latency of 3 words.}
    \label{fig:intro}
\end{figure}

Simultaneous machine translation (SMT) aims to produce translations based on partial input from the source language, enabling real-time communication across language barriers~\cite{gu-etal-2017-learning}. Despite its potential, current SMT methods often struggle to maintain high translation quality while achieving low latency. As shown in Figure \ref{fig:intro}, the translation quality of the previous state-of-the-art method SM2~\cite{yu-etal-2024-self} drops quickly as the latency decreases. The major source of such quality drop is insufficient source information.

To deal with insufficient information, human interpreters develop techniques to anticipate future source input to reduce interpretation lag~\cite{seeber2001intonation}. Human interpreters often predict upcoming words, such as nouns and verbs, based on context, language structure, prior knowledge, familiarity with the topic, etc. Most existing SMT models ignore this technique or only implicitly use it~\cite{ma-etal-2019-stacl,Ma2020Monotonic,zhang-feng-2022-information,miao-etal-2021-generative,10446517}.

A large language model (LLM) can predict the continuation of a source sentence given its prefix, mimicking this human-like anticipation~\cite{NEURIPS2020_1457c0d6,touvron2023llama2openfoundation}. However, perfectly predicting the continuation of a sentence is challenging due to the versatility of human language. For instance, given the prefix "The cat is chasing", various completions like "a mouse", "a bird", or "a dog" are all plausible. This variability makes it essential to develop a mechanism that ensures the translation remains consistent with the actual source when relying on the predictions.

In this work, we design a novel SMT policy, \textbf{T}ranslation by \textbf{A}nticipating \textbf{F}uture (\method), that achieves high-quality translation at an extremely low latency.  \method~utilizes an LLM to predict multiple possible continuations of the source input and translates each one with an MT model. It then employs a majority voting mechanism to select the prefix agreed upon by most candidates, ensuring consistency with source input. \method~works on any combinations of pretrained MT models and LLMs without the need for further finetuning and can be further generalized to existing SMT policies. Experimental results demonstrate that \method~consistently improves quality-latency trade-off over existing methods by up to 5 BLEU score at a latency of 3 words on four language directions. We conduct an in-depth analysis of the behavior of \method~to figure out the impact of LLM predictions. Finally, we find that providing the LLM with longer context further reduces the latency without sacrificing the translation quality. 

%% file: latex/sections/2_related_works.tex
\section{Related Works}

Recent advances in simultaneous machine translation majorly focus on its policy, either rule-based or learned adaptive. 
Rule-based policies include Wait-$K$ and its variants~\cite{ma-etal-2019-stacl,zeng-etal-2021-realtrans,elbayad20_interspeech}, Local Agreement (LA)~\cite{local_agreement}, Hold-$N$~\cite{liu20s_interspeech}, RALCP~\cite{ralcp} etc. 
Wait-$K$ waits for $K$ tokens at the beginning and then alternates between Read and Write actions. 
LA generates a full hypothesis at each step and writes the longest common prefix (LCP) of recent hypotheses. 
Hold-$N$ removes the last $N$ tokens of the full hypothesis and writes the rest. 
RALCP generates multiple hypotheses with beam search and outputs a common prefix that most hypotheses agree upon. 

Though rule-based policies are easy to implement, learned adaptive policies demonstrate better quality-latency trade-offs. 
\citet{Ma2020Monotonic} proposed monotonic multihead attention to model the policy.
\citet{miao-etal-2021-generative} developed a generative framework with a latent variable to make decisions. 
\citet{zhang-feng-2022-information} quantified the information weight transported from source to target and made decisions based on the amount of information received. 
\citet{zhang2023hidden} modeled the simultaneous translation process as a hidden Markov model and optimized the likelihood of target sequence over multiple steps. 
\citet{10446517} further bridged the gap between SMT models and offline MT models. 
\citet{yu-etal-2024-self} is the state-of-the-art adaptive policy that resolves the insufficient exploration issue of prior works by individually optimizing each source-target state.

These learned adaptive policies implicitly model the future source input through prefix-to-prefix training. \citet{DBLP:conf/icassp/YinZLTZ24} proposes to use a language model to explicitly predict one more source token and rewrite the translation if the prediction is wrong. However, rewriting makes it non-monotonic which limits its application to speech-to-speech scenarios.
\method~explicitly models the future source with an LLM so that it is not restricted by the limited MT training data, and maintains the best translation quality at a low latency without rewriting using majority vote. Also, \method~requires no model training and works on any combination of MT models and LLMs, thus easy to implement.

%% file: latex/sections/3_method.tex
\section{Method}

\begin{figure*}
    \centering
    \includegraphics[width=0.9\linewidth]{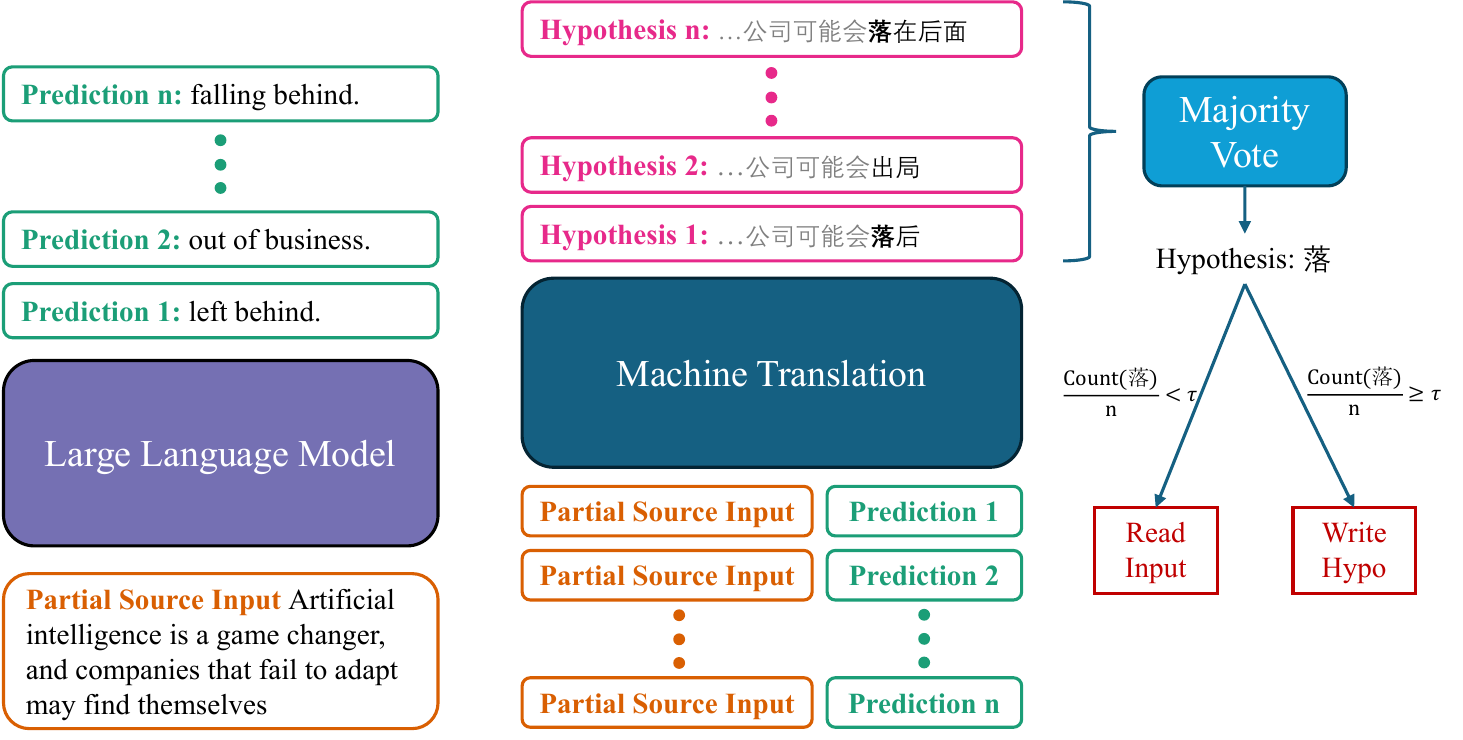}
    \caption{
    An overview of \method. \method~utilizes a large language model to predict multiple possible future continuations based on partial source input, and each continuation is translated using a machine translation model. Finally, \method~applies a majority voting mechanism to select the most agreed-upon hypothesis. The system commits to the translation if the frequency of the selected hypothesis exceeds a threshold $\tau$.}
    \label{fig:method}
\end{figure*}

\subsection{Problem Formulation}

Define $\vx_{1:j}=(x_1,\cdots,x_j)$ as partial text input and $\vy_{1:i} = (y_1,\cdots,y_i)$ as partial hypothesis that is already generated. Define delay $g_i$ as the number of source tokens read when generating $y_i$. Let $\pi(\vx_{1:j}, \vy_{1:i})\in [0,1]$ be the policy given $\vx_{1:j}$ and $\vy_{1:i}$, where $\pi(\vx_{1:j}, \vy_{1:i})$ is the probability to write $y_{i+1}\sim P_{MT}(y_{i+1}~|~\vx_j,\vy_{i})$ with $g_{i+1}=j$, and $1 - \pi(\vx_{1:j}, \vy_{1:i})$ is the probability to read $x_{j+1}$. The complete simultaneous translation process can be formulated as follows,
\begin{align}
    P(\vy,\vg|\vx) &= \prod_{i=1}^{|\vy|} P_{MT}(y_i|\vx_{1:g_i},\vy_{1:i-1})\times \nonumber\\    
    \pi(\vx_{1:g_i},&\vy_{1:i-1})\prod_{j=g_{i-1}}^{g_i-1}(1-\pi(\vx_{j},\vy_{i-1})). 
\end{align}

Once the translation is finished, we evaluate its quality and latency, respectively. The quality is assessed by comparing $\vy$ to the ground-truth $\vy^*$ using a metric $Q(\vy,\vy^*)$. The latency is assessed using delay $\vg=(g_1,g_2,\cdots)$ together with $\vx$, $\vy$ and $\vy^*$ using a latency function $L(\vx,\vy,\vy^*, \vg)$.

\subsection{Translation by Anticipating Future}

Define $P^*_{LM}(x_{j+1}~|~\vx_{1:j})$ as the ground-truth distribution of input, i.e., the oracle language model. Let $\vx_{1:j}$ and $\vy_{1:i}$ be the partial input and the generated hypothesis. An oracle simultaneous translation model $P^*_{MT}$ generates translation $y_{i+1}$ as if it knows the ground-truth input distribution, i.e.,
\begin{align}
    P_{MT}^*(y_{i+1}~|~\vx_{1:j},\vy_{1:i}) = & \nonumber \\
    \E_{\vx_{j+1:}\sim P^*_{LM}}[ P_{MT}^*(y_{i+1}~&|~\vx_{1:j},\vx_{j+1:},\vy_{1:i}) ] \label{eq:oracle}
\end{align}
where $\vx_{j+1:}$ is sampled from the oracle language model $P^*_{LM}(\cdot~|~\vx_{1:j})$.

One approach to approximating such an oracle model is to train the output distribution given partial input to be close to that given full input. However, given the limited MT training data compared with that used for language model pre-training, it is hard for the MT model to translate while anticipating future input. 

We propose to separate the translation and the anticipation. The MT model focuses only on the translation, while the language model handles the anticipation. The overview of our method is shown in Figure \ref{fig:method}. Since we cannot access the ground-truth language model $P^*_{LM}$, we approximate it with a pre-trained language model $P_{LM}$. Then we sample $n$ continuations of length $l$ from $P_{LM}$, 
\begin{align}
    \vx^1_{j+1:j+l},\cdots,\vx^n_{j+1:j+l} \label{eq:1st_step}
\end{align}
where $x^t_{j+r} \sim P_{LM}(~|~\vx_{1:j},\vx^t_{j+1:j+r-1})$ for $t\in[1,n]$ and $r\in[1,l]$. 
Note that we do not need to sample infinitely long continuations here because distant future source inputs will less likely affect the next token of the hypothesis.

Once we have the sampled continuations, we concatenate the received source input with the continuations and obtain the output distribution of the MT model for each of them
\begin{align}
    P^t_{MT} = P_{MT}(\cdot~|~\vx_{1:j},\vx^t_{j+1:j+l},\vy_{1:i}).
\end{align}
Finally, we aggregate the $n$ distributions $P^1_{MT},\cdots,P^n_{MT}$ using an aggregation function $f$ and obtain the translation $y_{i+1}$ at this step,
\begin{align}
    y_{i+1} = f(P^1_{MT},\cdots,P^n_{MT}). 
\end{align}
Inspired by RALCP~\cite{ralcp}, let $h^t = \arg\max P^t_{MT}$ and we design the aggregation function $f$ to be 
\begin{align}
    f(P^1_{MT},\cdots,P^n_{MT}) = \text{Majority}(h^t_1,\cdots,h_n^t) \label{eq:hyp}
\end{align}
where  $\text{Majority}()$ outputs the most common one of all inputs.

\begin{figure*}[t]
    \centering
    \includegraphics[width=\linewidth]{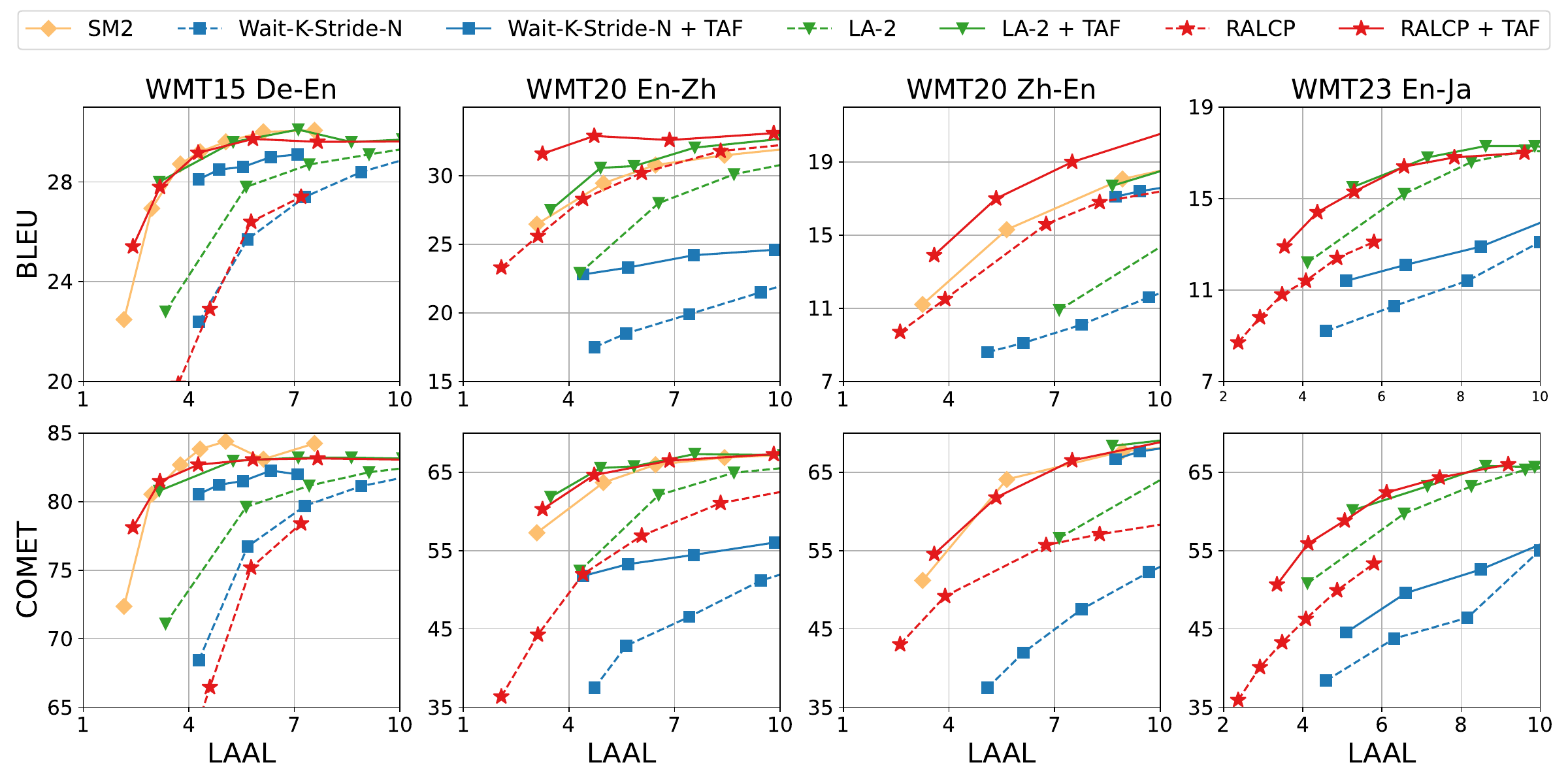}
    \caption{The quality-latency trade-off of \method~and other baselines. The quality is evaluated with both BLEU and COMET and the latency is evaluated with length adaptive average lagging (LAAL). \method~consistently improves existing policies and RALCP + \method~achieves the best performance across four language directions, with at most 5 BLEU scores improvement at a latency of 3 words in En-Zh direction.}
    \label{fig:main}
\end{figure*}

\subsection{Policy}

The policy of \method~is also a function of the output distributions,
\begin{align}
    \pi(\vx_{1:j},\vy_{1:i}) = \frac{1}{n}\text{Count}(f(P^1_{MT},\cdots,P^n_{MT}))
    \label{eq:policy}
\end{align}
where $\text{Count}(f(P^1_{MT},\cdots,P^n_{MT}))$ is the number of occurrences of the most common output. 
Intuitively, if $P^1_{MT},\cdots,P^n_{MT}$ are vastly different from each other, then it is unlikely there will be a definite output at this step, thus we should choose to read more input. Otherwise, if most distributions are close to each other, then we are confident there will be a definite output and should output the one with which most distributions agree. 

During inference, we select a threshold $\tau\in[0,1]$ and decide to write if $\pi(\vx_{1:j},\vy_{1:i})\geq \tau$ and read otherwise. We can then obtain a quality-latency trade-off by adjusting this threshold. 

We can also use other existing policies. We only need to switch the policy function $\pi$ to those existing ones. For example, when using Wait-$K$ policy~\cite{ma-etal-2019-stacl}, we wait for $K$ source tokens at the beginning, then generate one token at each step using Equation \ref{eq:1st_step}-\ref{eq:hyp}.

%% file: latex/sections/4_experiment.tex
\section{Experiment Setups}

\begin{figure*}[ht]
    \centering
    \includegraphics[width=0.85\linewidth]{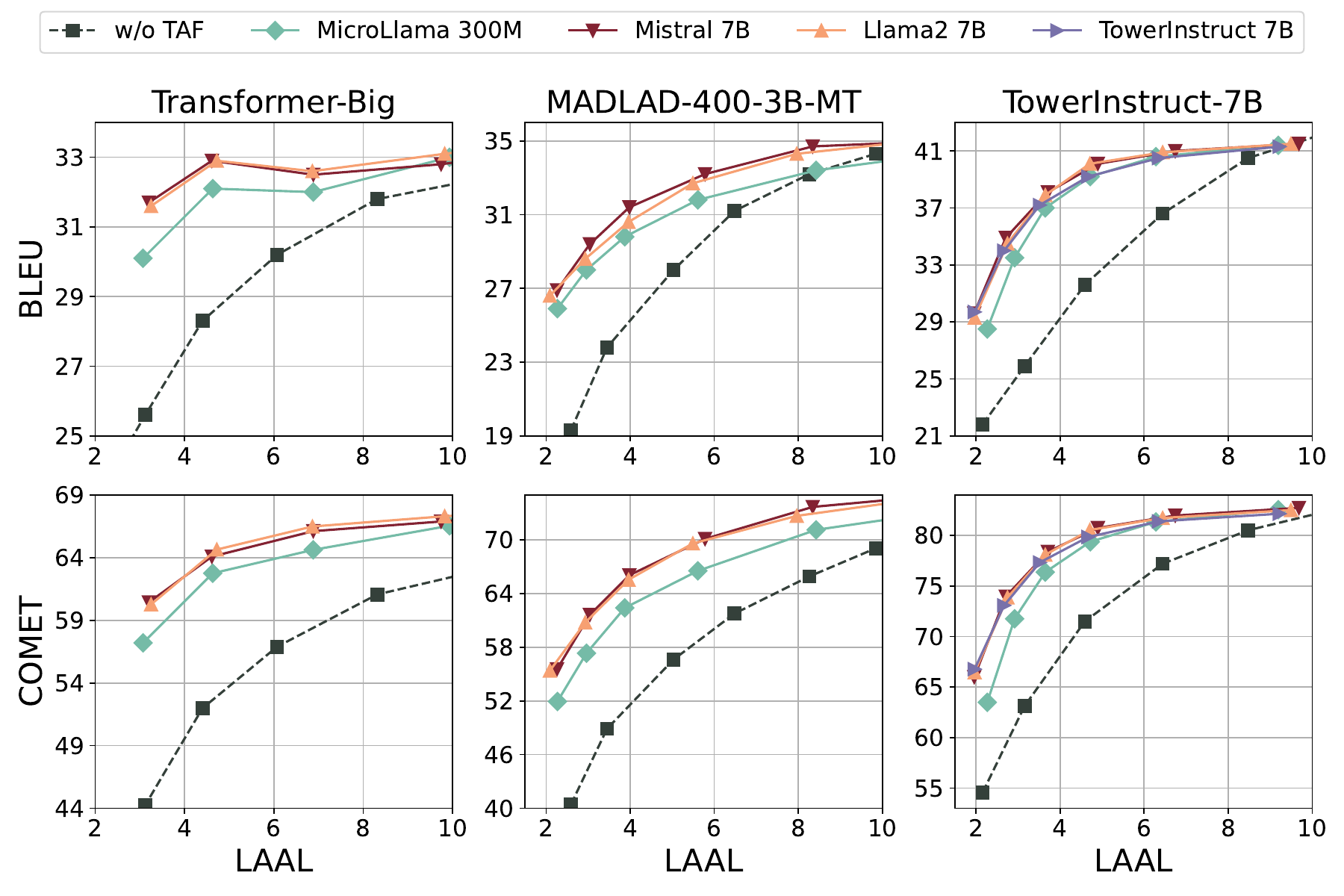}
    \caption{The quality-latency trade-off of \method~with different MT models and LLMs combinations. We also include the base RALCP (w/o \method) results as a reference. \method~is universally effective on all combinations with at least 5 BLEU score improvement at a latency of 3 words.}
    \label{fig:different_models}
\end{figure*}

\subsection{Datasets}

We use the following datasets to train the offline MT model and evaluate different systems. 

\paragraph{De-En} We use WMT15 as the training set which contains 4.5M sentence pairs, newstest2013 as the validation set, and newstest2015 as the test set. 

\paragraph{En-Zh/Zh-En} We use WMT20 as the training set which contains around 40M sentence pairs, newstest2019 as the validation set, and newstest2020 as the test set.

\paragraph{En-Ja} We use WMT23 as the training set which contains around 29M sentence pairs, newstest2020 as the validation set, and newstest2023 as the test set.

\subsection{System Settings}

\paragraph{\method} 
We default use Llama 2 7B base~\cite{touvron2023llama2openfoundation} as the language model for future prediction. We sample $n=10$ future continuations with top-$k$ sampling using $k=10$. The maximum length of each continuation is $l=10$. We sweep the threshold $\tau$ from $0.5$ to $1.0$ with step size $0.1$. We use Transformer-Big~\cite{NIPS2017_3f5ee243} as the architecture of our offline machine translation model for all language directions. The training details are reported in Appendix \ref{apdx:omt_training}.

\paragraph{Wait-$K$-Stride-$N$}\citep{zeng-etal-2021-realtrans} waits $K$ words at the beginning and then alternate between generating $N$ words and reading one more word. We enumerate $K$ in $[3,6,9,12,15]$. $N=1$ for De-En/Zh-En and $N=3$ for En-Zh/En-Ja. 

\paragraph{Local Agreement $N$ (LA-$N$)}\citep{local_agreement} generates a hypothesis after reading each word and outputs the longest common prefix of the last $N$ hypotheses. We vary the source segment size (the number of tokens read at each step) from 1 to 5. 

\paragraph{RALCP}\citep{ralcp} outputs a relaxed longest common prefix (LCP) of the beam search candidates after reading each word and finishing the beam search generation. Unlike LCP, where all candidates agree with the prefix, the relaxed prefix is the prefix that at least a fraction of candidates agree with. We vary the fraction from 0.1 to 1.0 with step size 0.1. The beam width is 40 for De-En/En-Zh and 10 for Zh-En/En-Ja.

Since \method~is compatible with existing policies, we also evaluate \method~with the above three policies. Note that applying \method~on top of RALCP is equivalent to the policy mentioned in Equation \ref{eq:policy}, and we make sure the product of the number of candidates $n$ and the beam width is equal to the beam width used in RALCP to have fair comparison.

\paragraph{SM2} Additionally, we compare our method with the state-of-the-art learned adaptive SMT method SM2~\citep{yu-etal-2024-self}. SM2 individually optimizes the decision at each step and uses prefix sampling to ensure sufficient exploration during training. For the SM2 baseline, we follow the instructions in its original paper to obtain models of similar size to our offline model.

\subsection{Evaluation Metrics}

We use SimulEval~\citep{ma-etal-2020-simuleval} to evaluate our method and other baselines. We evaluate the quality of translation by comparing the hypothesis with the ground-truth translation using BLEU~\cite{papineni-etal-2002-bleu} and COMET \cite{10.1162/tacl_a_00683}\footnote{\url{https://huggingface.co/Unbabel/XCOMET-XXL}}. The latency is evaluated with Length Adaptive Average Lagging (LAAL)~\citep{papi-etal-2022-generation}. Note that we treat the word (En, De) or character (Zh, Ja) as the unit during evaluation instead of the BPE token, following the practice of recent IWSLT competitions~\cite{ahmad-etal-2024-findings}. The latency calculated with the word or character is more intelligible than that computed with the BPE token and also enables fair comparison of models with different tokenizations. 

\section{Results}

\subsection{Best Translation Quality at an Extremely Low Latency Across Language Directions}

We evaluate whether \method~improves the quality-latency trade-off over existing policies and compare it with the state-of-the-art learned model SM2.
As shown in Figure \ref{fig:main}, \method~consistently improves Wait-$K$-Stride-$N$, LA-2, and RALCP on all language directions for at least 6 BLEU scores when the latency is around 2 words. Among them, RALCP with \method~is showing the best results. It is competitive with SM2 in the De-En direction and outperforms SM2 in the En-Zh and Zh-En directions with at most 5 BLEU scores at the latency of 3 words. COMET shows similar results as BLEU. These results demonstrate that \method~achieves the state-of-the-art translation quality at an extremely low latency across different language directions. 

\subsection{Generalizable to Existing Pretrained MT models and LLMs}
\method~does not require sophisticated finetuning for simultaneous translation, making it easily generalizable to other pre-trained MT models and LLMs. 
Here we show that \method~is effective across different combinations of such models. 

For MT models, we choose MADLAD-400-3B-MT~\citep{kudugunta2023madlad} and TowerInstruct 7B~\citep{alves2024tower}, as the former is a typical encoder-decoder translation model and the latter is a typical decoder-only translation model. For language models, we examine MicroLlama-300M\footnote{\url{https://huggingface.co/keeeeenw/MicroLlama}} and Mistral-7B-v0.3~\cite{jiang2023mistral7b} to find out whether \method~works for smaller LMs and other LLMs of similar size. Besides, since TowerInstruct-7B itself is an LLM, we also include the result using it for both prediction and translation. We conduct experiments on WMT20 En-Zh with RALCP+\method~policy since it performs the best, as shown previously.

Results are shown in Figure \ref{fig:different_models}. \method~is universally effective on all combinations of MT models and LLMs. At a latency of 3 words, \method~improves more than 5 BLEU scores using Transformer-Big and MADLAD-400-3B-MT and more than 8 BLEU scores using TowerInstruct-7B. Also, we observe that using Mistral and Llama2 shows similar performance. A smaller LLM like MicroLlama leads to a performance drop of less than 2 BLEU scores but still 4 BLEU scores better than the base RALCP policy.

We also demonstrate the perplexity of LLMs on the source text in Table \ref{tab:perplexity} together with their latency and quality scores. The effectiveness of \method~is closely correlated with the LLM perplexity on the source text.

\begin{table}[t]
    \centering
    \begin{tabular}{c|c|c|c}\toprule
        MT Model & Size & w/o \method & w/ TAF \\\midrule
        Transformer-Big & 0.2B & 72ms & 1067ms \\
        MADLAD-400 & 3B & 383ms & 1339ms \\
        TowerInstruct & 7B & 905ms & 1506ms \\\bottomrule
    \end{tabular}
    \caption{The average wall-clock time per step with and without \method~on different MT backbones. LLM is fixed to 7B. As the size of the MT backbone increases, the relative additional overhead introduced by LLM prediction decreases.}
    \label{tab:computation}
\end{table}

\subsection{Computation Cost}
\label{sec:comp}

\method~introduces additional computation overhead to conventional SMT. We report the average wall-clock time of generating a full hypothesis with RALCP+\method~versus the base RALCP policy on different combinations of MT models and a 7B LLM. We run experiments on an A6000 GPU and an AMD EPYC 9354 32-Core CPU. The results are shown in Table \ref{tab:computation}. When using a small MT model Transformer-Big (0.2B), 7B LLM introduces significant computational overhead. As the size of the MT model gets larger, the relative overhead reduces.


\subsection{Sensitivity to Hyperparameters}

\begin{figure}[t]
    \centering
    \includegraphics[width=0.7\linewidth]{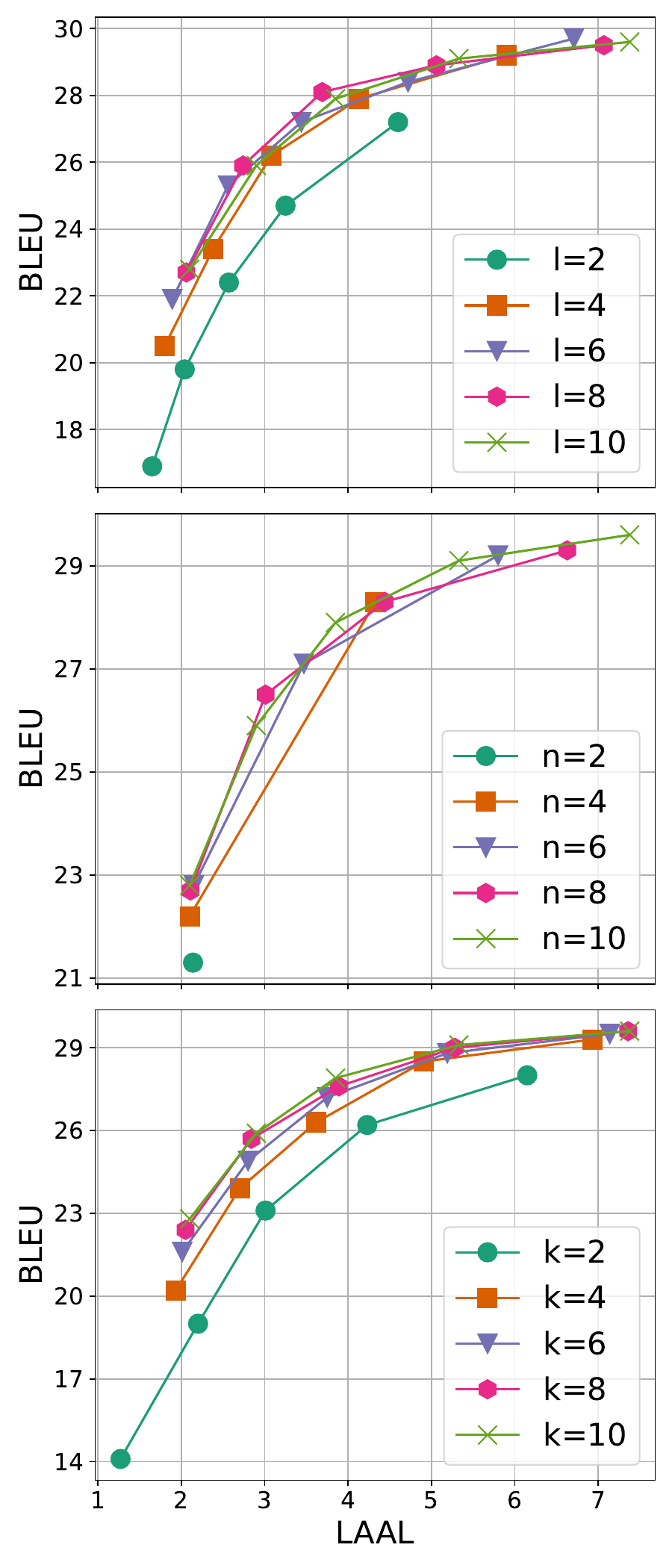}
    \caption{The quality-latency trade-off of \method~with different hyperparameters on WMT15 De-En. The improvement from \method~saturates after $l\geq 6$, $n\geq 4$ and $k\geq 8$, which means we can further reduce the computation overhead of \method.}
    \label{fig:hyperparameters}
\end{figure}

\begin{figure*}[ht]
    \centering
    \includegraphics[width=1.0\linewidth]{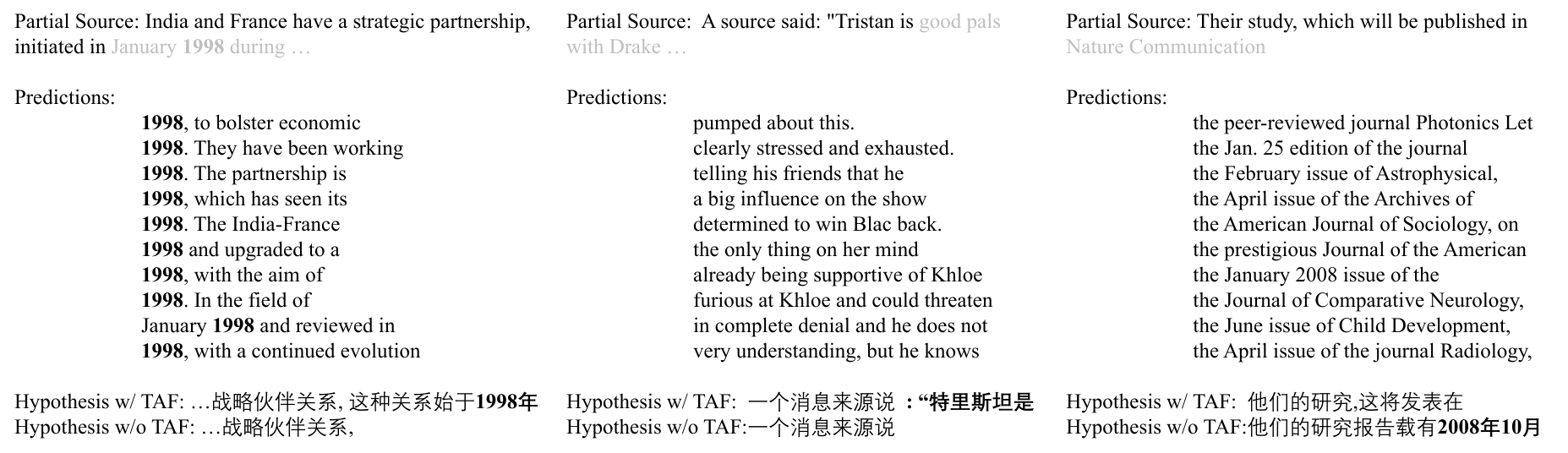}
    \caption{Cases of three typical patterns where \method~improves the quality-latency trade-off. The greyed-out text is ground-truth future source text not yet read. The first case demonstrates \method~reduces the latency with correct predictions. The second case shows \method~reduces the latency with additional context. The last case illustrates \method~reduces hallucination with additional context.}
    \label{fig:case123}
\end{figure*}

\begin{figure}[ht]
    \centering
    \includegraphics[width=0.7\linewidth]{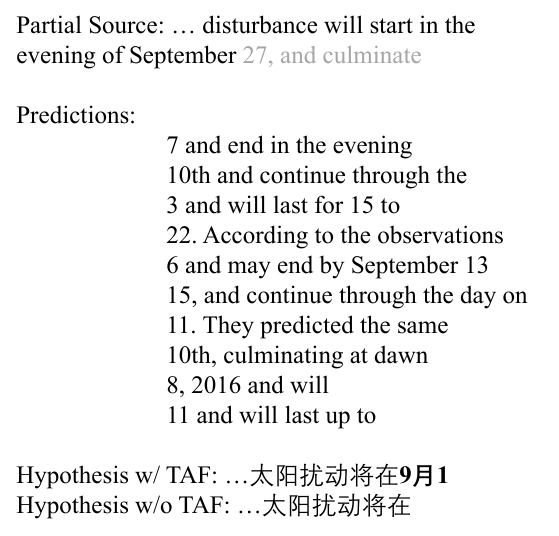}
    \caption{Case where \method~introduces additional hallucination due to LLM bias.}
    \label{fig:case45}
\end{figure}

We examine how \method~is sensitive to the choice of hyperparameters. We evaluate on De-En direction and use RALCP+\method. We vary the number of tokens to predict $l$, the number of continuations $n$, and the $k$ for top-$k$ sampling from $[2,4,6,8,10]$. We sweep $\tau$ from $0.5$ to $0.9$ with step size $0.1$ for each configuration. Note that for $n=2$ and $\tau<1$, there is only one data point since RALCP always follows the first of $n=2$ translations.

Results are shown in Figure \ref{fig:hyperparameters}. As the sampling length $l$ increases, the quality-latency trade-off improves and saturates after $l\geq 6$. This confirms that more distant source text will have a smaller impact on the hypothesis generation. As the number of candidates $n$ increases, the trade-off also gets better but quickly saturates after $n\geq 4$, which means we can further reduce the computation cost without sacrificing too much performance. Another observation is that a larger number of candidates with the same threshold $\tau$ will have a larger latency since it requires more randomly sampled candidates to agree with each other. As $k$ increases, we also see better results, which is consistent with Equation \ref{eq:oracle} since we want the sampled distribution to be as close to the oracle distribution as possible. 



%% file: latex/sections/5_analysis.tex
\section{How \method~Impacts the Translation}
\label{sec:analysis}


\begin{table*}[t]
    \centering
    \begin{tabular}{c|c|c|c} \toprule
        Pattern & Frequency & $\Delta$ COMET & $\Delta$ LAAL \\ \midrule
        All 100 Instances & 100\% & +6.78 & -0.25 \\ \midrule
        $\downarrow$ Latency w/ Correct Prediction & 44\% & +10.61 & -0.45 \\
        $\downarrow$ Latency w/ Additional Context & 53\% & +5.42 & -1.05 \\
        $\downarrow$ Hallucination w/ Additional Context & 39\% & +24.44 & +0.29 \\ \midrule
        $\uparrow$ Hallucination w/ LLM Bias & 28\% & -6.99 & -0.31\\\bottomrule
    \end{tabular}
    \caption{Statistics on the impact of \method~on hypothesis generation across different patterns in 100 manually examined instances. $\Delta$ represents the difference between the results of RALCP with and without \method.}
    \label{tab:pattern}
\end{table*}

We manually examine 100 instances in the En-Zh direction with the Llama2 as the LLM and MADLAD-400-3B-MT as the MT model. We compare the trajectory between RALCP with and without \method~using $\tau=0.6$. We find four major patterns where \method~improves the quality-latency trade-off or hurts it. We show the frequency of these patterns in Table \ref{tab:pattern}. If a pattern occurs multiple times in an instance, we count it once. 

\paragraph{Reduce Latency with Correct Prediction (+)} When LLM predicts the correct future words of the source, the MT model can translate those words before they appear and reduce the latency. This happens often for those entity words. Since the WMT data is from news, LLM can guess the right entity with high probability given enough context. We illustrate this with an example in Figure \ref{fig:case123}. "India and France have a strategic partnership in 1998" is already known by LLM, so the MT model can translate the year of this partnership before it is read. 

\paragraph{Reduce Latency with Additional Context (+)} When LLM prediction is not correct, it still provides additional context for the policy so that it is more confident to generate a translation of what is already read. This is illustrated in the second example in Figure \ref{fig:case123}. With LLM prediction, the MT model is confident in translating "Tristan is" into Chinese, but without future prediction, the model stays conservative and needs more input to continue the translation. 

\paragraph{Prevent Hallucination with Additional Context (+)} The MT model will often generate hallucinated content that does not appear in the source or generate low-quality translation with insufficient context. The prediction of LLM provides additional context to the policy to realize the possible continuations of the source and stays conservative if the current context is not enough for generation. This is illustrated in the last example in Figure \ref{fig:case123}. Without future prediction, the MT model generates hallucination (highlighted in bold) without any such information from the source. This is probably caused by bias during MT model training and can be avoided with extended context from the LLM. 

\paragraph{Introduce Additional Hallucination (-)} In certain cases, the bias of LLM also introduces additional hallucination during simultaneous translation. As shown in Figure \ref{fig:case45}, the LLM predicts "September 1" when only "September" appears in the source, but the true date of the event is "September 27". As shown in Table \ref{tab:pattern}, it slightly worsens translation quality in some cases. However, when considering both the reduction and the introduction of hallucinations, \method~ultimately improves overall translation quality.


\section{Better Prediction with Longer Context}
\label{sec:long_context}

\begin{figure}[t]
    \centering
    \includegraphics[width=\linewidth]{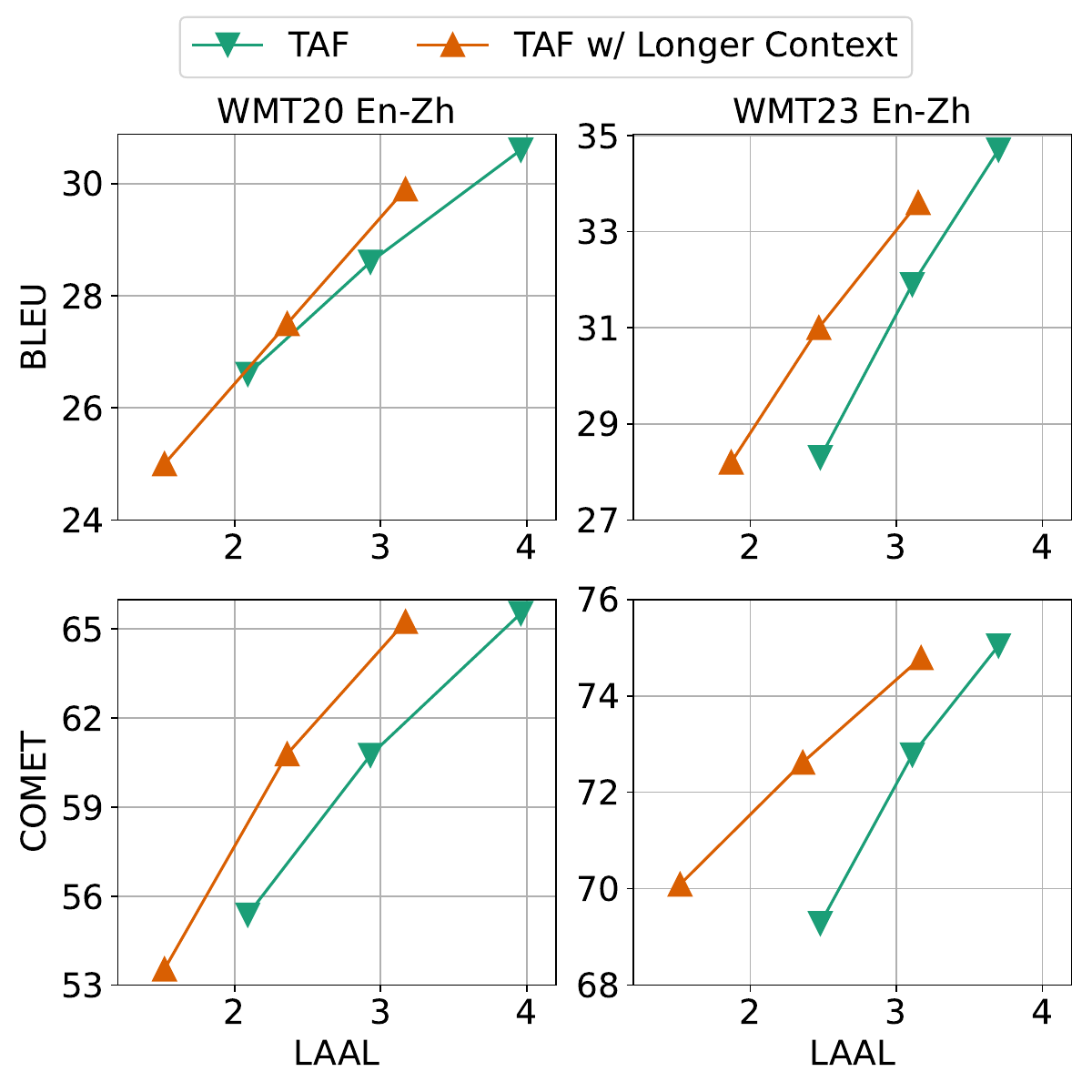}
    \caption{The quality-latency trade-off of \method~with and without longer context on WMT20 and WMT23 En-Zh test set. Longer context consistently reduces the latency of \method~by 0.5 to 1 word.}
    \label{fig:longer_context}
\end{figure}

In real-world SMT applications, such as multilingual conferences, speeches can last minutes or hours, allowing models to leverage the full context of prior speech. However, WMT datasets are pre-segmented into sentences. When evaluating SMT on isolated sentences, LLMs cannot access the prior context of the document, making it challenging to anticipate future content accurately.

For instance, given the partial source "Regular physical activity can significantly reduce the risk of", the next phrase could refer to various diseases related to physical activity. However, if the prior context focuses on heart health, the continuation will likely be "heart disease".

To better simulate real-world scenarios, we prepend the current partial source with previous sentences from the same document, allowing the LLM to predict based on this extended context. Before passing these predictions to the MT model, we remove the past sentences to ensure that translation remains sentence-level. This setup guarantees that the LLM's context is the only variable being tested.


Results are shown in Figure \ref{fig:longer_context}. \method, when using a longer context, consistently reduces latency by 0.5 to 1 word while maintaining comparable translation quality on both WMT20 and WMT23 En-Zh. Since WMT23 test data is less likely to have appeared in Llama2’s training set, the improvement from \method~may not be solely due to triggering the LLM's memory. Instead, the longer context could enable the LLM to make more informed predictions about future content.

%% file: latex/sections/6_conclusion.tex
\section{Conclusion}

We propose \method~, a novel SMT method that translates by anticipating future source input. Experiments on four language directions show that \method~achieves state-of-the-art quality-latency trade-off and is universally effective on different combinations of pretrained MT models and LLMs. Our manual analysis reveals how \method~impacts the translation output. Finally, we show that \method~can be further improved with a longer context.

%% file: latex/sections/limitations.tex
\section*{Limitations}

Apart from the additional computation cost mentioned in Section \ref{sec:comp} and the hallucination caused by LLM bias in Section \ref{sec:analysis}, we have yet to explore other choices of the aggregation function. It can be simply a mean pooling function or a more advanced function that takes the semantic meaning into account. Besides, our experiments focus on X-En and En-X directions. X-X directions are not covered yet. Also, our experiments are only on text-to-text translation. The major obstacle to migrating \method~to speech-to-text translation is that predicting continuous future audio signals is very hard. A possible solution could be a cascade speech-to-text model with \method~applied to the transcribed speech.

%% file: latex/sections/appendix.tex
\begin{table*}[ht]
\centering
\begin{tabular}{llrrr}
\toprule
\textbf{Model} & \textbf{Context} & \textbf{Perplexity} & \textbf{LAAL} & \textbf{COMET} \\
\midrule
MicroLlama-300m & Sentence & 113.63 & 2.27 & 51.93 \\
Llama2-7b       & Sentence & 12.86  & 2.09 & 55.37 \\
Mistral-7b      & Sentence & 14.52  & 2.26 & 55.54 \\
Llama2-7b       & Document & 5.76   & 2.36 & 60.79 \\
\bottomrule
\end{tabular}
\caption{LLM perplexity correlates well with simultaneous translation performance.}
\label{tab:perplexity}
\end{table*}

\section{Training Details of Offline MT}
\label{apdx:omt_training}

We trained our NMT models (Transformer, $12 \times 6$ layers, $d_\text{model}=1024$, $d_\text{inner}=4096$, $n_\text{heads}=16$) with Adam optimizer and inverse square root annealing~\cite{NIPS2017_3f5ee243} with $7.5$K warmup steps and a maximum learning rate of $10^{-3}$. The models were trained for a maximum of $100$K steps with a dropout of $0.1$ on intermediate activations and label smoothing with $\alpha=0.2$. Our De$\rightarrow$En models used joint BPE vocabulary of $16384$ tokens and En$\leftrightarrow$Zh/Ja used separate vocabularies with the same number of tokens per language.

\section{LLM Perplexity on Source Text}

To find out the correlation between LLM’s next token prediction and the final simultaneous translation performance, we measure the perplexity of MicroLlama-300m, Llama2-7b and Mistral-7b on source sentences of WMT20 En-Zh test set. We also measure the perplexity of Llama2-7b on unsegmented source documents as in Section \ref{sec:long_context}. The results are shown in Table \ref{tab:perplexity}.